\documentclass[letterpaper, 10 pt, conference]{ieeeconf}  

\IEEEoverridecommandlockouts                              

\usepackage{algorithmic}
\usepackage{graphicx}
\usepackage{textcomp}
\usepackage{xcolor}
\usepackage{caption}
\usepackage{ragged2e}
\usepackage{mathtools}
\usepackage{array}
\usepackage{booktabs}
\usepackage{hyperref}
\usepackage{subcaption}
\usepackage{xcolor}  
\usepackage{amsmath} 
\usepackage{amsfonts} 
\usepackage{bm} 

\usepackage{booktabs}
\usepackage{multirow}

\usepackage{todonotes}
\usepackage{comment}

\title{\LARGE \bf
Multimodal Transformers for Real-Time Surgical Activity Prediction 
}

\author{Keshara Weerasinghe$^{**1}$, Seyed Hamid Reza Roodabeh$^{**1}$,  Kay Hutchinson$^{1}$,  Homa Alemzadeh$^{1}$ 
\thanks{*This work was supported in part by the National Science Foundation under Grants CNS-2146295 and DGE-1842490.}
\thanks{** denotes equal contribution.}
\thanks{$^{1}$Department of Electrical and Computer Engineering, University of Virginia, Charlottesville, VA 22903 USA.
        {\tt\small \{cjh9fw,ydq9ag,kch4fk,ha4d\}@virginia.edu}.} 
        \thanks{The code for this paper is available at \url{https://github.com/UVA-DSA/MTRSAP}}
}

\begin{document}

\maketitle
\thispagestyle{empty}
\pagestyle{empty}

\begin{abstract}

Real-time recognition and prediction of surgical activities are fundamental to advancing safety and autonomy in robot-assisted surgery. This paper presents a multimodal transformer architecture for real-time recognition and prediction of surgical gestures and trajectories based on short segments of kinematic and video data. 
We conduct an ablation study to evaluate the impact of fusing different input modalities and their representations on gesture recognition and prediction performance.
We perform an end-to-end assessment of the proposed architecture using the JHU-ISI Gesture and Skill Assessment Working Set (JIGSAWS) dataset. Our model outperforms the state-of-the-art (SOTA) with 89.5\% accuracy for gesture prediction through effective fusion of kinematic features with spatial and contextual video features. It achieves the real-time performance of 1.1-1.3ms for processing a 1-second input window by relying on a computationally efficient model. 

\end{abstract}

\section{INTRODUCTION}

Surgical robots translate the intricate movements of a surgeon's hands, wrists, and fingers into precise actions performed by miniature surgical instruments and offer many advantages, including improved visual perception, heightened surgical dexterity, reduced incision size \cite{kumar2016minimally}, and shortened postoperative recovery periods \cite{finkelstein2010open}. Their adoption in surgical specialties such as urology, gynecology, and general surgery is not only enhancing surgical precision and quality, but is also opening avenues for the development of autonomous systems \cite{bonne2022digital, gonzalez2023asap, han2022systematic, maier2017surgical}, automated skill assessment \cite{fard2018automated, reiley2010motion, zia2018automated}, and error detection \cite{yasar2019context, yasar2020real,li2022runtime}. However, the development of these systems in robot-assisted minimally invasive surgery (RMIS) requires the understanding and perception of surgical activities carried out during surgical operations to support lower-level analysis \cite{van2021gesture,hutchinson2023evaluating} of procedures. 

Multiple modalities of data are available from surgical robots including video and kinematic data that can be used separately or in combination for the recognition and prediction of surgical gestures.
Previous works have proposed methods for the recognition of gestures based on kinematic data from the surgical robot \cite{shi2022recognition}, \cite{farha2019ms}, \cite{dipietro2019segmenting}, \cite{menegozzo2019surgical}, \cite{van2020multi} or video data of the surgical scene \cite{lea2016temporal}, \cite{zhang2020symmetric}, \cite{funke2019using}, \cite{sarikaya2019surgical}. More recently, there have been several efforts that utilize the fusion of kinematic and video data for gesture recognition~\cite{qin2020temporal}, \cite{van2022gesture}. 
However, most prior works have concentrated on the recognition of gestures based on the observation of a complete trial of a surgical task as opposed to short temporal segments observed at runtime. This limitation hinders the practical implementation of runtime gesture recognition in realistic environments for applications such as safety monitoring \cite{yasar2019context,yasar2020real,li2022runtime}, training \cite{despinoy2015unsupervised}, autonomy~\cite{autonomy, ginesi2020autonomous}, and teleoperation~\cite{gonzalez2023asap,bonne2022digital}. Recognizing a surgical gesture within a short temporal segment (e.g., a 1-second window) can help with timely intervention and feedback during simulated or real surgical tasks. However, window-based gesture recognition and prediction are more challenging compared to analyzing the entirety of a surgical trial due to the significantly reduced availability of contextual information. 
To our knowledge, fusion of multiple data modalities and the impact of different representations of modalities for window-based gesture recognition and prediction have not been studied before.
Moreover, the \textit{end-to-end} performance evaluation of gesture recognition and prediction, which is important for real-time interventions in real-world deployment, has not been explored in previous works.

To address these challenges, we propose methods that utilize the rich information embedded in different modalities by fusing kinematic and video data and exploring 
different representations of these modalities on the end-to-end performance of surgical activity (gesture and trajectory) prediction. 

The main contributions of the paper are as follows:
\begin{itemize}
    \item We propose a multimodal transformer model that utilizes the fusion of different modalities to recognize surgical gestures based on short temporal segments of surgical activity data (1-second), which is then used to predict surgical gestures and trajectories for a short temporal segment in the future.
    \item We conduct an ablation study on the impact of different modalities (including video and robot kinematics) and different representations of certain modalities (e.g., features extracted using ResNet50 \cite{he2016deep}, Spatial CNN \cite{lea2016segmental}, and contextual representations \cite{hutchinson2023towards} of video data) on gesture recognition and prediction.
    \item We perform an end-to-end evaluation of our proposed model on the publicly available JIGSAWS dataset and show that our model can outperform a previous transformer model \cite{shi2022recognition} in gesture prediction and trajectory prediction with a prediction accuracy of \textbf{89.5\%} vs. 84.6\%, while achieving a real-time performance of, respectively, 1.3ms and 1.1ms for a 1-second window in gesture recognition and prediction. 
\end{itemize}

\section{PRELIMINARIES}

\subsection{Surgical Gestures and Context}

Previous works have modeled surgical procedures using a hierarchy \cite{neumuth2011modeling} of steps, phases, tasks, gestures, and low-level motion primitives~\cite{hutchinson2023towards}.
Gestures are 
defined as purposeful actions imbued with semantic content that are specific and often involve 
particular instruments or objects. \cite{hutchinson2023compass} proposed a framework that further decomposes surgical gestures into a sequence of elementary instrument movements referred to as ``motion primitives'', which encompass basic actions such as pushing, pulling, and grasping. 
\cite{hutchinson2023compass} defined ``context'' as a set of states that describe the status and interactions among surgical tools, objects, and the physical environment which can be inferred from video data \cite{hutchinson2023towards, li2023robotic}.
A change in context happens as the result of the execution of a motion primitive within a gesture or task~\cite{hutchinson2023compass}. 
In this paper, we leverage context as an alternative representation of the information within the video to recognize surgical activities.


\subsection{JIGSAWS Dataset}

JIGSAWS~\cite{gao2014jhu} is a publicly available dataset of surgical tasks performed using the da Vinci robot~\cite{dimaio2011vinci}. It includes synchronized kinematic, video, and gesture transcripts collected from executions of three fundamental surgical tasks on a bench-top model by eight surgeons of three expertise levels. We evaluate our methods using 39 trials of the Suturing task.

The kinematic data in JIGSAWS captures the Cartesian positions (\(\mathbf {p} \in \mathbb{R}^3\)), rotation matrices (\(\mathbf{R} \in \mathbb{R}^{3 \times 3}\)), linear velocities (\(\mathbf{v} \in \mathbb{R}^3\)), rotational velocities (\(\boldsymbol{\omega} \in \mathbb{R}^3\)), and grasper angles (\(\theta\)) for left and right tools 
for both patient-side manipulators (PSM) and master-side manipulators (MTM), resulting in a total of 76 features sampled at 30Hz. The video data is collected at 
30fps from an endoscopic camera. The dataset also contains manual annotations for gestures based on a predefined surgical activity vocabulary (see Table 2 in \cite{gao2014jhu}), along with the skill levels of the subjects.

\subsection{Gesture Recognition}

Gesture recognition plays a vital role in identifying the present state of surgical procedures, enabling the detection of safety violations~\cite{yasar2019context,yasar2020real, li2022runtime} and facilitating the prediction of future surgical actions \cite{davinciNet, shi2022recognition} with enhanced accuracy and confidence. Early research on surgical gesture recognition relied on probabilistic graphical models like Hidden Markov Models \cite{murphy2004towards, van2021gesture}, whereas contemporary studies predominantly focus on the utilization of deep learning (DL) techniques \cite{lea2016temporal, shi2022recognition} based on video \cite{lea2016segmental},\cite{sarikaya2019surgical},\cite{funke2019using},\cite{zhang2020symmetric} and/or kinematic \cite{gurcan2019surgical},\cite{dipietro2019segmenting},\cite{shi2022recognition}, \cite{van2020multi} data.
Specifically, Temporal Convolutional Networks (TCNs) \cite{lea2016segmental, lea2016temporal, lea2017temporal} have been shown to efficiently capture temporal information for action segmentation based on video data from the JIGSAWS dataset.
\cite{van2022gesture} employs the TCN in a parallel two-stream network with weighted fusion, and \cite{qin2020temporal} utilizes TCN and LSTM to leverage multimodal data in improving surgical gesture recognition, but does not address gesture and trajectory prediction. 
Although both works also consider real-time implementations of their models, in general, limited attention has been directed toward real-time recognition based on short temporal segments through multimodal fusion. 
Furthermore, a gap remains in understanding the significance of different representations of certain modalities and their impact on gesture recognition accuracy and real-time performance as it is imperative for developing systems such as online safety monitoring \cite{yasar2020real,li2022runtime} for RMIS.

\begin{figure*}[t!]
      \includegraphics[width=\textwidth]{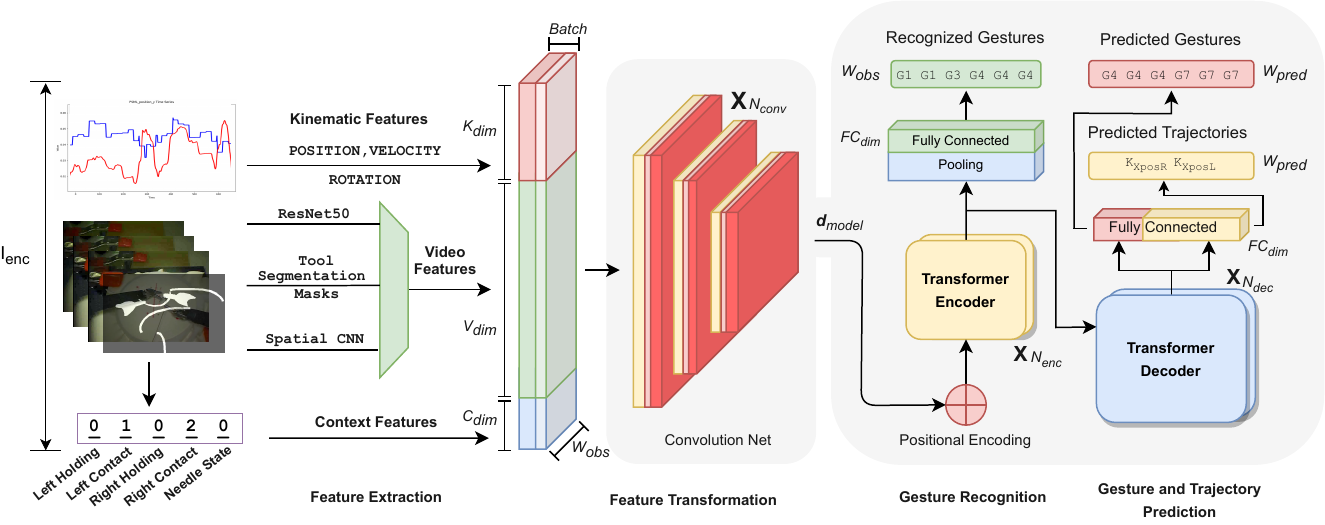}
      \caption{Overall Architecture for End-to-End Real-Time Surgical Activity Recognition and Prediction}
      \label{model}
      \vspace{-1.5em}
\end{figure*}

\subsection{Gesture and Trajectory Prediction}
Prediction of surgical activities including gestures and tool trajectories is an area of growing interest and applicability such as in visual window \cite{visualtrack1, visualtrack2} and surgical instrument \cite{insttrack1} tracking. The precise and timely predictions of surgical states and trajectories can improve the success rates of tele-operated surgical procedures \cite{sarters, digitaltwin} and guide the surgical tools in real-time autonomous operations~\cite{autonomy, ginesi2020autonomous}.

While some earlier studies relied on conventional methods, such as silhouette-based instrument tracking using Kalman filters \cite{insttrack1}, recent research is leveraging DL techniques. For instance, daVinciNet~\cite{davinciNet} adopts a fusion approach involving encoded multimodal features using LSTM \cite{lstm} 
along with feature and temporal attention mechanisms. Similarly, \cite{shi2022recognition} proposes a pipeline of three consecutive transformer models \cite{vaswani2017attention} for gesture recognition, gesture prediction and trajectory prediction based on kinematic features. 

\section{METHODS}
In this section, we introduce our model for gesture recognition, prediction, and trajectory prediction within the context of short temporal segments. This model is built upon an adaptation of the original transformer model proposed by \cite{vaswani2017attention} for Natural Language Processing (NLP) and incorporates the fusion of multimodal data. The transformer model has proven its excellence in NLP tasks due to its capabilities of identifying long-term patterns from historical data \cite{wen2022transformers}, \cite{zerveas2021transformer}, \cite{liu2021gated}, \cite{devlin2018bert} as well as in the domain of sequence generation \cite{zhou2021informer}, \cite{radford2018improving}. We aim to utilize the strengths of the transformer architecture and introduce changes that can adapt to the context of RMIS for runtime surgical gesture recognition, prediction, and trajectory prediction. We also evaluate the impact of different modalities and their representations on the recognition and prediction performance.

Our proposed model is structured as a three-part pipeline, including the stages for Feature Extraction and Transformation, Gesture Recognition, and Gesture/Trajectory Prediction, as illustrated in Figure \ref{model}. 
More specifically, we process the input data features for an observation window, spanning from \(t + 1\) to \(t + W_{obs}\), recognize the gesture(s) being performed in that window, and use this output along with other features to predict future gesture(s) and trajectory coordinates for a prediction window \(t + W_{obs} + 1\) to \(t + W_{obs} + W_{pred}\). 

The Gesture Recognition stage consists of a transformer encoder whose output is fed as input to another transformer model, separately trained for Gesture/Trajectory Prediction. We utilize multi-task learning to train a single transformer model to simultaneously predict both gestures and trajectories for a prediction window \(W_{pred}\). This unified end-to-end model architecture is different from previous works \cite{shi2022recognition} and \cite{davinciNet}, and particularly well-suited for real-time tasks, such as error detection and recovery~\cite{yasar2020real}, where timely response and reduced computational complexity are needed.


\subsection{Feature Extraction and Transformation}
The first stage of the pipeline transforms the multimodal input data into features representing the rich information embedded in the data. 
Of the 76 kinematic features in the JIGSAWS dataset, we explore using subsets of features including just the 38 kinematic features from the PSM side (\(K_{38}\)) and only the 14 kinematic features representing the position, velocity, and gripper angles from PSM (\(K_{14}\)).

In transfer learning, pre-trained CNNs are widely used to extract latent features from raw images as input for downstream tasks. We utilize different SOTA methods to extract feature representations from video data,  including \(V_{Res}\) extracted using pre-trained ResNet50 \cite{he2016deep}, \(V_{Spatial}\) extracted using a Spatial CNN proposed for action segmentation~\cite{lea2016segmental}, and \(V_{Seg}\) surgical instrument and object segmentation masks extracted using memory networks, which have been shown to be effective in capturing interactions between instruments and objects~\cite{li2023robotic}. Since the raw image features of the segmentation masks are high dimensional and the actual objects only occupy a relatively small area of the image, we first resize each segmentation frame by a factor of ten and then apply Principal Component Analysis (PCA) \cite{pca} to the resized images to extract a more compact representation of each frame.
We also use the surgical context \cite{hutchinson2023compass} defined as a state vector \(C\) representing the interactions between surgical instruments (e.g., graspers, scissors) and objects (e.g., needle, thread) in the surgical scene, which has been proposed as a fine-grained representation of surgical activity~\cite{hutchinson2023compass,hutchinson2023towards}. 
and can be inferred from video data using a combination of knowledge and data-driven methods~\cite{li2023robotic,hutchinson2023towards}. 

We perform an ablation study on the above set of features to assess the effects of different modalities and their respective representations on gesture recognition and prediction. This study aims to determine the most suitable fusion of features for the subsequent feature transformation stage.

We draw upon the insights presented in \cite{lea2016temporal} for the transformation of selected features from the kinematic and video data, which are then given as input to our model. We adopt the encoder component of the TCN model in \cite{lea2017temporal} to efficiently capture features from the fused inputs as shown in Figure \ref{model}. The TCN encoder employs a stack of hierarchical \(N_{conv}=3\) temporal convolutional layers, pooling, and non-linear activations which effectively capture robust temporal relationships while enhancing computational efficiency, as shown in \cite{lea2017temporal}.  We modify the final convolutional filter to output a feature vector of dimension \(d_{model}\) that is subsequently fed into the encoder component of the transformer model for gesture recognition. During gesture and trajectory prediction, the feature transformation stage is skipped, and input features are directly fed into the encoder.

\begin{figure*}[t!]
      \includegraphics[width=\textwidth]{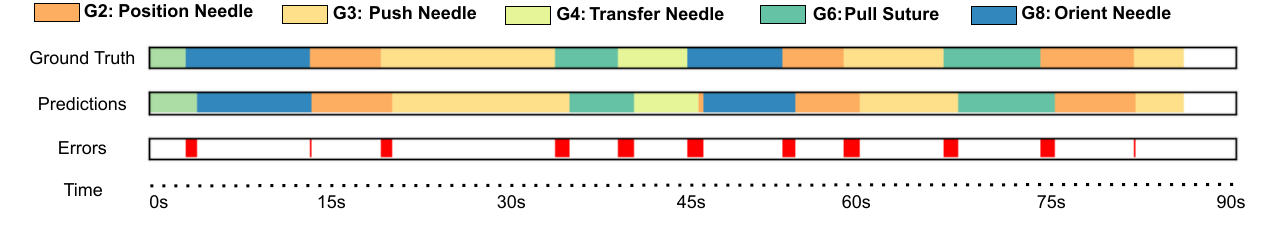}
      \vspace{-2em}
      \caption{A sample timeline of a Suturing trial, illustrating the gestures executed throughout the trial. Top Row: Actual gestures, Middle Row: Predicted gestures, Bottom Row: Error intervals (often occurring when transitioning to the next gestures). 
      }
      \label{predbarplot}
      \vspace{-1.5em}
\end{figure*}

\subsection{Gesture Recognition}

For runtime gesture recognition, we 
leverage the encoder component derived from the transformer encoder-decoder \cite{vaswani2017attention} architecture. In NLP, the encoder module of the transformer architecture 
performs the comprehension and extraction of 
information embedded within the input text \cite{devlin2018bert} and is mainly used for classification tasks.
We regard the recognition of surgical gestures based on time series data as an adaptation of transformers for the time series classification, as evident in 
\cite{liu2021gated} and \cite{zerveas2021transformer}. We also employ the decoder module of the transformer architecture to predict gestures and trajectories, as these tasks entail generative aspects that align well with the decoder's capabilities. 

In the classical transformer implementation \cite{vaswani2017attention}, an embedding layer transforms the input data into sequential token embeddings before sending it to the encoder. This step is not needed in surgical gesture recognition where numerical data is used instead of textual input. Our encoder uses a multi-headed attention mechanism to generate a feature vector of dimension \(d_{model}\).
A fully connected layer \(FC_{dim}=10\) is appended to the output of the encoder as depicted in Figure \ref{model} to attain the desired dimension for the gesture output vector \(O_{enc}\), representing each gesture class. 

Real-time gesture recognition is achieved by using an observation window \(W_{obs}\) with a duration of 30 samples (corresponding to \(1 s\)) representing a short temporal segment as the input to the model. The output consists of gesture labels at runtime of length \(W_{obs}\) (see Figure~\ref{predbarplot}). A tumbling window approach \cite{patroumpas2006window}, instead of a sliding window, is used to decrease computational overhead.




\subsection{Gesture and Trajectory Prediction} 
Our integrated framework for simultaneous gesture and trajectory prediction hinges on the recognition module's output to accurately generate gesture labels for the observation window \(W_{obs}\). These labels, coupled with the encoder's output, and the original input features prior to the feature extraction stage, form the input to the transformer decoder. The decoder comprises \(N_{dec}\) layers, with each layer utilizing \(H_{dec}\) attention heads. This design allows the model to effectively consider various aspects of the input data and observed gestures when predicting the correct gesture at each time step. The hidden dimensions of both the transformer encoder and decoder are identical.

The decoder's output undergoes two linear transformations to produce the final outputs. One transformation maps the output to a set of probability weights for the gesture prediction task, while the other maps the output to the 3D Cartesian trajectory coordinates of the two robot end-effectors, represented by six real-valued numbers (\(X, Y, Z\) coordinates for each end-effector). 
We adopted a cumulative \(L_{2}\) function over the prediction window to compute the regression loss for trajectory prediction. This loss is calculated based on the differences between the ground truth and predicted trajectory variables. 
We employed categorical cross-entropy over the prediction window for our classification loss 
which measures the disparity between the ground truth and predicted gesture labels. The ultimate loss function is a weighted combination of these two terms. 
The weights are hyper-parameters that are fine-tuned to achieve optimal model performance between trajectory prediction and gesture prediction.

\begin{table*}[t!]
\caption{Performance of gesture recognition in the ablation study of input features for Suturing task:
\(K_{38}\) = 38 Kinematic features, 
\(K_{14}\) = 14 Kinematic features, 
\(C\) = Surgical Context features \cite{hutchinson2023compass},
\(V_{Spatial}\) = Video features from Spatial CNN \cite{lea2016segmental},
\(V_{Res}\) = Video features from ResNet50 \cite{he2016deep},
\(V_{Seg}\) = Video features from Tool Segmentation Masks \cite{li2023robotic}
}
\centering
\begin{tabular}{lcccccc}

\toprule
\multicolumn{1}{c}{\multirow{2}{*}{\begin{tabular}[c]{@{}c@{}}Input \\ Features\end{tabular}}} & \multirow{2}{*}{Accuracy (\%)} & \multicolumn{1}{c}{\multirow{2}{*}{\begin{tabular}[c]{@{}c@{}}Edit\\ Score (\%)\end{tabular}}} & \multirow{2}{*}{F1@10 (\%)} & \multirow{2}{*}{F1@25 (\%)} & \multirow{2}{*}{F1@50 (\%)} & \multicolumn{1}{c}{\multirow{2}{*}{\begin{tabular}[c]{@{}c@{}}Inference\\ Time (ms)\end{tabular}}} \\
\multicolumn{1}{c}{}    &  & \multicolumn{1}{c}{}     &    &        &        & \multicolumn{1}{c}{}                \\ \midrule
\(K_{38}\) & 71.1 & 66.8 & 69.0 & 68.0 & 60.4 & 0.60 \\
\(K_{14}\) & 74.8 & 72.3 & 72.5 & 71.3 & 64.6 & \textbf{0.55 }\\
\(C\) & 74.3 & 71.3 & 75.6 & 74.2 & 65.6 & 0.57 \\
\(V_{Spatial}\) & \textbf{80.7} & \textbf{81.2} & \textbf{82.0 }& \textbf{80.8} & \textbf{72.3} & 1.52 \\
\(V_{Res}\)  & 69.8 & 66.6 & 70.5 & 68.8 & 58.9 & 1.67 \\
\(V_{Seg}\) & 47.7 & 45.2 & 47.6 & 44.5 & 33.6 & 1.45 \\
\cmidrule(lr){1-7}
\(K_{14}\) + \(C\) & 78.6 & 76.2 & 77.8 & 76.5 & 70.0 & \textbf{0.58} \\
    \(K_{14}\) +\(V_{Spatial}\)  & 83.5 & \textbf{84.0} & 86.3 & 85.8 & 79.0 & 1.14 \\
\(K_{14}\) + \(V_{Res}\) & 76.2 & 71.4 & 76.1 & 75.0 & 66.8 & 1.64 \\
\(K_{14}\) + \(V_{Seg}\) & 57.5 & 55.0 & 59.0 & 57.8 & 46.8 & 1.14 \\
\(C\) + \(V_{Spatial}\) & \textbf{84.4} & 83.4 & \textbf{86.5} & \textbf{86.1} & \textbf{80.6} & 1.24 \\
\cmidrule(lr){1-7}
\textbf{\(K_{14}\)+ \(C\) + \(V_{Spatial}\)} & \textbf{87.1} & \textbf{83.9} & \textbf{87.3} & \textbf{86.5} & \textbf{81.1} & 1.32 \\
\(K_{14}\) + \(V_{Seg}\) + \(C\) & 71.3 & 69.3 & 73.4 & 72.3 & 65.4 & 1.33 \\
\(K_{14}\) + \(V_{Seg}\) + \(C\) +  \(V_{Spatial}\)  & 71.5 & 69.1 & 72.7 & 71.8 & 64.9 & \textbf{1.31} \\
  \bottomrule
\end{tabular}%
\label{ablation}
\end{table*}

\section{EXPERIMENTAL EVALUATION}

\subsection{Experimental Setup}

The experiments were done on a PC with an Intel Core i9-12900K 3.20GHz, 32GB RAM, and an NVIDIA GeForce RTX 3080 Ti 12GB GPU running Ubuntu 20.04.6 LTS.

We used the Leave-One-User-Out (LOUO) \cite{ahmidi2017dataset} cross-validation strategy to evaluate the model performance and to conduct the ablation study of the impact of different modalities and their representations. For surgical gesture recognition, we used the original sampling rate of data at \(30Hz\) with \(W_{obs} = 30\) samples which is equivalent to \(1s\). For gesture prediction and trajectory prediction, we used the same temporal window length of 1 second, but downsampled the data from \(30Hz\) to \(10Hz\), thus having \(W_{pred} = 10\).

Across all input configurations, for gesture recognition we maintained model hyper-parameters at \(N_{enc}=3\), \(H_{enc}=2\), and \(d_{model}=60\), whereas for gesture and trajectory prediction we maintained \(N_{dec}=2\) and \(H_{dec}=4\). Model training was done using the Adam optimizer \cite{adam} with a dynamic learning rate, as outlined in \cite{vaswani2017attention}. The training duration spanned 20 epochs with a batch size of 10.

\subsection{Metrics}

We evaluate the performance of each module and the overall end-to-end pipeline using the following metrics.

\textbf{Gesture Recognition and Prediction:} We use the standard accuracy and edit score metrics~\cite{yujian2007normalized} to compare the recognized/predicted gestures to the ground truth labels. We evaluate the window-based classification using the F1@X metric~\cite{lea2017temporal}, which defines recognized/predicted windows that overlap with actual windows by more than ``X" percent as true positives and those with less overlap as false positives.



\textbf{Trajectory Prediction:} In order to assess the performance of the trajectory prediction module, we analyze the difference between the ground truth and predicted left and right end-effector trajectories within the Cartesian endoscopic reference frame using the standard metrics of Root Mean Square Error (RMSE), Mean Absolute Error (MAE), and Mean Absolute Percentage Error (MAPE).

\begin{table*}[]
\centering
\caption{Performance of gesture prediction in the ablation study of input features for Suturing task:
\(G\) = Surgical Gestures for the observation window. Other notations are the same as Table I.
}
\vspace{-0.5em}
\resizebox{\textwidth}{!}{
\begin{tabular}{lccccccc}
\toprule
\multicolumn{1}{c}{\multirow{2}{*}{\begin{tabular}[c]{@{}c@{}}Input\\ Features\end{tabular}}} & \multicolumn{3}{c}{Ground truth Gestures} & \multicolumn{3}{c}{Recognized Gestures} & \multirow{2}{*}{\begin{tabular}[c]{@{}c@{}}Inference\\ Time (ms)\end{tabular}} \\ \cmidrule(lr){2-4} \cmidrule(lr){5-7}
 & Accuracy (\%)  & Edit Score (\%)  & F1 @ 10,25,50 (\%) & Accuracy (\%)  & Edit Score (\%)  & F1 @ 10,25,50 (\%) &  \\ \cmidrule(lr){1-8}   
 
\(K_{14}\) + \(G\)  & 85.4     & 87.0      & 81.3, 81.1, 78.5 & 80.1    & 82.8  & 78.1, 77.8, 77.1     & \textbf{0.49}  \\
\(K_{14}\) + \(G\)  + \(C\)  & 88.8 & 90.4  & 85.0, 83.9, 77.8  & 85.5 & 87.2  & 84.7, \textbf{84.4}, \textbf{82.9}  & 0.89  \\
\(K_{14}\) + \(G\)  + \(V_{Spatial}\) & \textbf{89.5} & \textbf{91.3}  & \textbf{87.8}, 84.4, 80.3 & 86.0    & 88.2    & \textbf{86.2}, 84.2, 80.5  & 1.08  \\
\(K_{14}\) + \(G\)  + \(V_{Spatial}\) + \(C\) & 87.1     & 90.7     & 86.6, \textbf{85.3}, \textbf{82.6} & \textbf{86.5}    & \textbf{89.8}     & 85.3, 82.3, 81.7  & 1.30    \\
\(K_{14}\) + \(G\)  + \(V_{Seg}\) & 86.6     & 88.8      & 81.1, 78.3, 77.0 & 83.3    & 84.9      & 80.3, 78.2, 74.0 & 1.19  \\ 
\midrule
daVinciNet \cite{davinciNet} & 84.3 & - & - & - & - & - & - \\
Transformer (MTMs) \cite{shi2022recognition} & 84.6 & - & - & - & - & - & - \\
Transformer (PSMs) \cite{shi2022recognition} & 84.0 & - & - & - & - & - & - \\
\bottomrule
\end{tabular}
}
\label{gpred}
\vspace{-1em}
\end{table*}

\textbf{Inference Time:} We report the inference times for different stages of the pipeline to assess the impact of different feature representations on real-time performance. The inference times are measured for a 1-second window, averaged across trials and subjects. The inference time for gesture recognition encompasses both feature extraction and recognition.

\subsection{Results}

\begin{table}[]
\centering
\caption{Performance of trajectory prediction in the end-to-end study of input features for Suturing task. RMSE and MAE are expressed in millimeters (mm):
}
\vspace{-0.5em}
\resizebox{\columnwidth}{!}{%
\begin{tabular}{@{}cccccccc@{}}
\toprule
Input Features & Metric & x1 & y1 & z1 & x2 & y2 & z2 \\
\midrule
\multirow{3}{*}{\(K_{14}\) + \(G\) } & RMSE & 5.3 & 4.5 & 5.9 & 6.16 & 6.37 & 6.74 \\
 & MAE & 4.79 & 4.1 & 4.87 & 5.75 & 6.04 & 6.09 \\
 & MAPE & 11 & 9.7 & 10.23 & 12.68 & 13.2 & 13.72 \\
 \cmidrule(l){1-8}
\multirow{3}{*}{\(K_{14}\) + \(G\)  + \(C\)} & RMSE & 5.11 & 4.32 & 5.8 & 5.34 & 5.56 & 5.65 \\
 & MAE & 4.43 & 3.77 & 4.68 & 4.23 & 4.34 & 4.45 \\
 & MAPE & 10.3 & 9.1 & 10.03 & 10.66 & 10.69 & 11.26 \\
  \cmidrule(l){1-8}
\multirow{3}{*}{\(K_{14}\) + \(G\)  + \(V_{Spatial}\)} & RMSE & 4.8 & \textbf{4.09} & 5.23 & \textbf{4.57} & \textbf{4.44} & \textbf{4.37} \\
 & MAE & 4.13 & \textbf{3.55} & 4.19 & \textbf{3.77} & \textbf{3.29} & \textbf{3.02} \\
 & MAPE & 7.6 & \textbf{8.4} & 9.45 & \textbf{8.52} & \textbf{9.06} & 9.74 \\
  \cmidrule(l){1-8}
\multirow{3}{*}{\(K_{14}\) + \(G\)  + \(V_{Spatial}\) + \(C\)} & RMSE & \textbf{4.75} & 4.14 & \textbf{5.17} & 5.2 & 4.6 & 4.41 \\
 & MAE & \textbf{3.91} & 3.6 & \textbf{3.99} & 3.8 & 3.73 & 4.1 \\
 & MAPE & \textbf{7.4} & 8.7 & \textbf{8.78} & 9.35 & 9.2 & \textbf{9.11} \\
  \cmidrule(l){1-8}
\multirow{3}{*}{\(K_{14}\) + \(G\)  + \(V_{Seg}\)} & RMSE & 4.92 & 4.49 & 5.76 & 5.4 & 4.9 & 4.55 \\
 & MAE & 4.3 & 4.02 & 4.61 & 4.16 & 3.96 & 4.12 \\
 & MAPE & 8.84 & 9.5 & 9.7 & 9.45 & 9.79 & 9.36 \\
\midrule
\multirow{2}{*}{daVinciNet \cite{davinciNet}} & RMSE & \textbf{2.53} & \textbf{1.89} & \textbf{2.96} & \textbf{3.15} & \textbf{3.5} & \textbf{3.91} \\
 & MAE & \textbf{2.07} & \textbf{1.51} & \textbf{2.46} & \textbf{2.78} & \textbf{3.06} & 3.50 \\
 & MAPE & \textbf{6.43} & \textbf{4.72} & \textbf{6.35} & \textbf{6.13} & \textbf{6.11} & \textbf{6.67} \\ 
 \cmidrule(l){1-8}
\multirow{2}{*}{Transformer \cite{shi2022recognition}} & RMSE & 3.15 & 3.03 & 3.30 & 3.93 & 4.21 & 4.22 \\
 & MAE & 2.86 & 2.85 & 3.00 & 3.60 & 3.88 & 3.84 \\
\bottomrule
\end{tabular}%
}
\label{tpred}
\end{table}

\begin{table}[]
\vspace{-0.5em}
\caption{Gesture recognition accuracy compared with related work for Suturing task under LOUO cross-validation
}
\vspace{-0.5em}
\resizebox{\columnwidth}{!}{%
\begin{tabular}{@{}ccccc@{}}
\toprule
 & Data Sources & Trial (\%) & 1s Window (\%) &  \\ \midrule
Transformer \cite{shi2022recognition} & PSM & \_ & 89.2 &  \\
Fusion-KVE \cite{qin2020temporal} & PSM + Video & \_ & 86.3 &  \\
MA-TCN \cite{van2022gesture} & PSM  + Video & 83.4 & \_ &  \\
TCN \cite{lea2016temporal} & Video & 81.4 & \_ &  \\
Our model & PSM + Video + Context & \textbf{86.1} & 87.3 &  \\
\bottomrule
\end{tabular}%
}
\label{table:gesturecomparison}
\end{table}

\subsubsection{Gesture Recognition}

Table \ref{ablation} shows the results of the ablation study on the impact of different modalities and their representations on the performance of runtime gesture recognition for the \textbf{Suturing} task. We observe that utilizing a subset of kinematic features, \(K_{14}\), yields superior results compared to utilizing all 38 kinematic features, \(K_{38}\). 
Thus, \(K_{14}\) represents the essential characteristics \cite{van2021gesture} of a surgical gesture while omitting non-significant kinematic variables that are typically associated with the Suturing task. These results suggest that utilizing a refined subset of kinematic features can lead to more precise and effective surgical gesture recognition with decreased computational overhead. 

Notably, the method of feature extraction from video as well as the fusion of video and kinematic features affect gesture recognition performance as shown in Table \ref{ablation}.
\(V_{Spatial}\) extracted using Spatial CNN \cite{lea2016segmental} contributes to significant improvements compared to using features extracted using ResNet50 \cite{he2016deep} due to the inherent capability of capturing spatio-temporal information from a video. Surgical context \(C\) improves performance by 5\% due to the ability of capturing contextual features highly specific for robotic surgery \cite{hutchinson2023compass} which signifies the temporal relationships of gesture sequences and transitions defined in a surgical process. We also observe that certain input feature combinations, such as \(K_{14}\) + \(V_{Seg}\) + \(C\) + \(V_{Spatial}\), can result in overfitting and ultimately lead to decreased performance of the model. 
We also observe that the expertise level of the subjects and the different approaches taken by the subjects to perform the same task impacts the recognition accuracy. Moreover, certain expert subjects maintain similar movements over multiple trials and novice subjects exhibit inconsistent movements. This also impacts the model's gesture recognition performance.

 As shown in Table \ref{table:gesturecomparison}, our model achieves comparable performance to SOTA gesture recognition models by demonstrating an accuracy of \textbf{87.3\%} for a 30-sample window (1s), while maintaining a low inference time of just \textbf{1.3ms}. This enables real-time performance by meeting the constraint of performing inference faster than the input data acquisition rate of 30Hz (33.33ms) \cite{quellec2014real}.  We note that the SOTA model presented by \cite{shi2022recognition} attains a slightly higher accuracy of 89.3\% for gesture recognition. The source code for \cite{shi2022recognition} was not available, so we were unable to reproduce their results.

\subsubsection{Gesture Prediction}
Table \ref{gpred} shows the results from our ablation and end-to-end experiments, which investigate the impact of different video feature representations on gesture prediction performance. When employing solely \(K_{14} + G\) as features, our results show agreement with prior works \cite{davinciNet}, \cite{shi2022recognition}. Notably, introducing context \(C\) as an additional feature led to major enhancements in performance, yielding an improvement of nearly \(3\%\) in accuracy and edit score. This underscores the significance of context features in enhancing the predictive capacity for future surgical activities. Furthermore, the inclusion of \(V_{Spatial}\) as a video data representation results in substantial performance improvements, leading to the combination \(K_{14} + G + V_{Spatial}\) outperforming the state-of-the-art with an accuracy of \textbf{89.5\%} and an edit score of \textbf{91.3\%}. Similar to gesture recognition, integrating both kinematic and video features yields superior prediction results. Although the utilization of \(V_{Seg}\) as a complement to kinematic features did yield improvements compared to baseline kinematic inputs, its impact was not as substantial as the spatio-temporal features or context variables. Additionally, the combination of three inputs \(K_{14} + G + V_{Spatial} + C\)  did not demonstrate significant performance enhancements and only increased the inference time of the prediction model. In summary, \(K_{14} + G + V_{Spatial}\) is the most effective set of features, providing the best performance while maintaining reasonable inference times. Figure \ref{predbarplot} shows the output of the gesture prediction over a sample suturing trial, using our top performing feature configuration.

End-to-end gesture prediction results are shown in Table \ref{gpred} where the output of the recognition model is used instead of the ground truth gesture labels from the observation window. Our best model outperforms previous work by about \(2.5\%\) in accuracy, and using \(K_{14} + G + V_{Spatial} + C\) seems to be more robust to inaccuracies in the gesture recognition outputs.

\begin{figure}[t!]
    \centering
     \vspace{-0.5em}
     \begin{subfigure}[b]{0.4\textwidth}
         \centering
         \includegraphics[width=\textwidth]{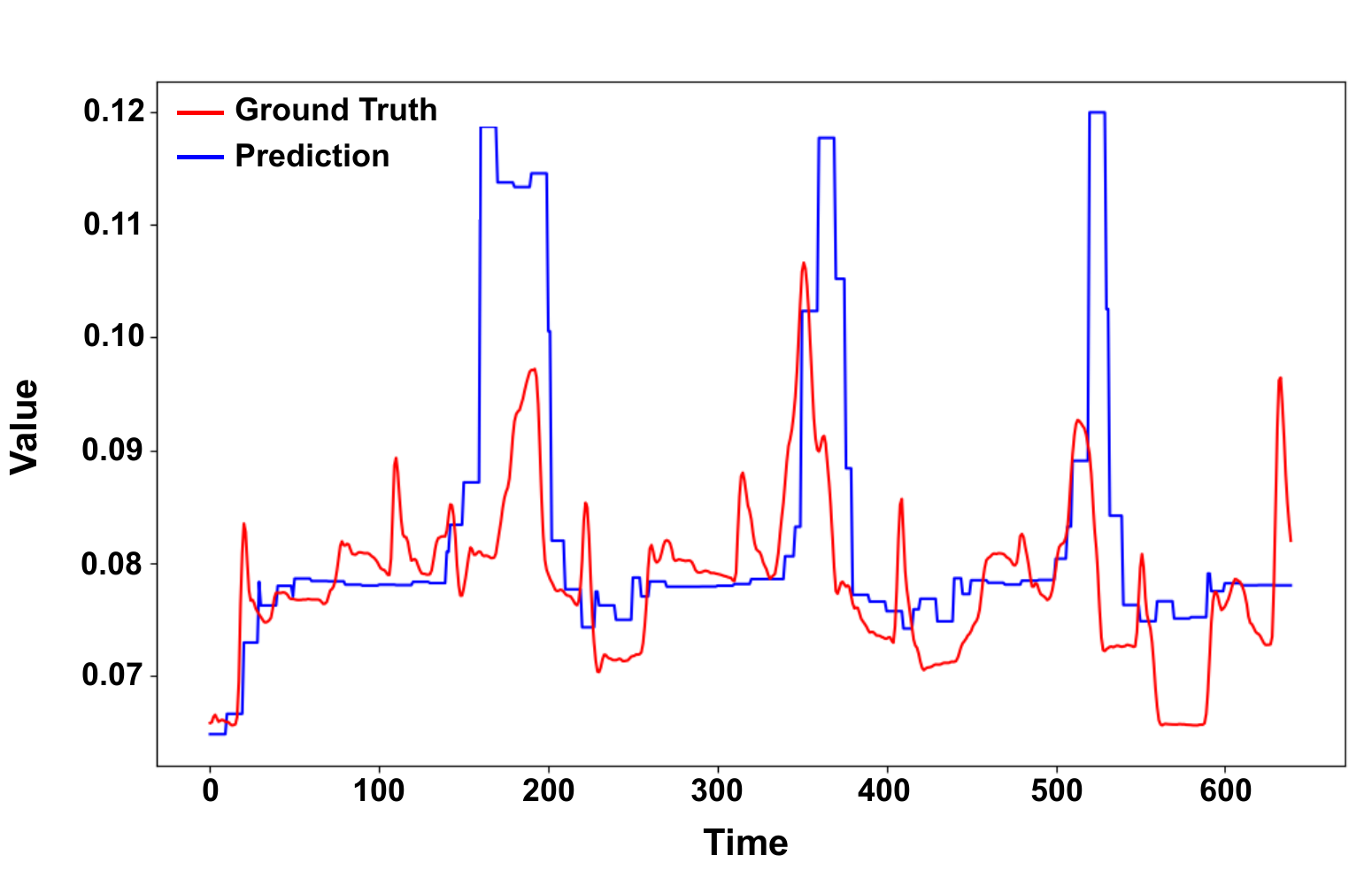}
         \caption{PSMR Position X}
         \label{fig:psmrx}
     \end{subfigure}
       \begin{subfigure}[b]{0.4\textwidth}
         \centering
         \includegraphics[width=\textwidth]{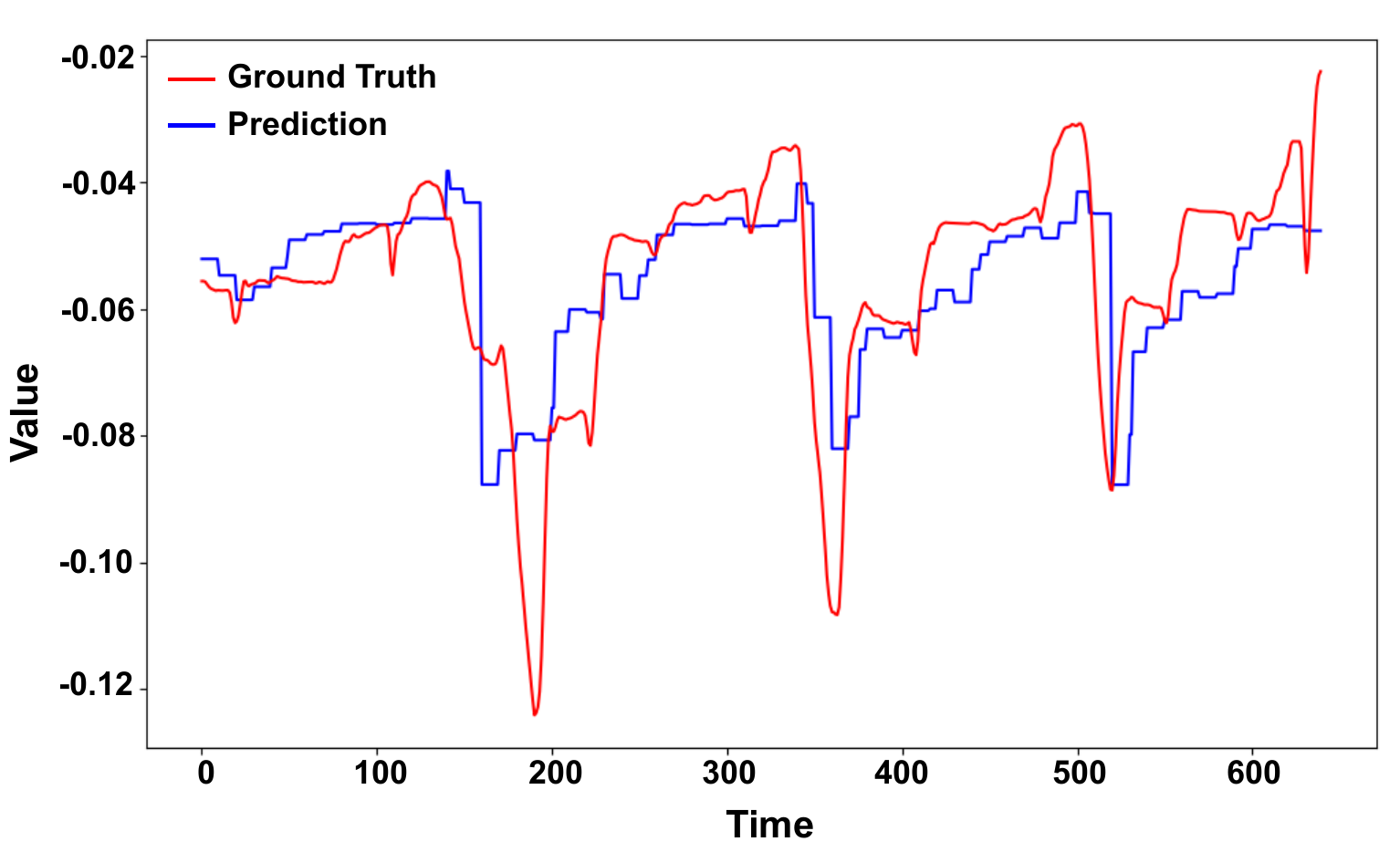}
         \caption{PSMR Position Z}
         \label{fig:psmrz}
     \end{subfigure}
    \caption{Trajectory Prediction results for X-axis and Z-axis position of the right instrument for a subject in the Suturing task}
    \label{fig:trajplot}
    
    \vspace{-2em}
\end{figure}
   
\subsubsection{Trajectory Prediction and End-to-End Evaluation}
Table \ref{tpred} shows end-to-end results for our trajectory prediction module. We ran the full model without the use of ground truth observation gestures (similar to the second column of the Table \ref{gpred}) and computed RMSE, MAE and MAPE metrics for Cartesian coordinates of each robot end-effector. In terms of the effectiveness of each input feature/video representation configuration in predicting accurate trajectories, the results are almost the same as gesture prediction. The \(K_{14} + G + V_{Spatial} + C\) configuration still performs best on average, although \(K_{14} + G + V_{Spatial} + C\) also appears to be comparable (e.g. for \(x1\) and \(z1\) coordinates). The inference times are also the same as gesture prediction since both predictions are generated simultaneously at the output of the transformer decoder. 
Figure \ref{fig:trajplot} shows the predicted \(x\) and \(z\) trajectories for the right patient-side grasper during a trial using our top performing feature configuration.
The Fundamentals of Laparoscopic Surgery guidelines \cite{FLS} recommend a tool trajectory error of up to 1 mm in Suturing. However, this is not achieved by SOTA, including the daVinciNet model~\cite{davinciNet}. Our approach using \(K_{14} + G + V_{Spatial} + C\) as input is not the most accurate, but it provides the best trade-off between accuracy and inference time. We observe performance fluctuations among different subjects, which appears to be a function of their level of expertise. The subjects with higher expertise have a better economy of motion and smoother trajectories than others. When learning on novices and intermediates and predicting for experts, we observe signs of over-compensation by the prediction model.

\section{CONCLUSIONS}
We presented a multimodal transformer architecture for real-time surgical gesture and trajectory prediction toward improving safety and autonomy in RMIS. 
This architecture outperforms the SOTA gesture prediction models by utilizing advanced video feature extraction techniques and achieves real-time performance by relying on a single transformer model.
We evaluated the efficacy of multiple input feature configurations for both the recognition and prediction tasks and the end-to-end pipeline. Our results indicate that the fusion of kinematic data with spatial and contextual video features consistently yields the best performance. 
Future work will focus on validating our proposed method using data collected from a wider range of surgical tasks, participants with a variety of surgical skills, and actual surgical procedures and on applying it to real-time safety monitoring.









\bibliographystyle{IEEEtran}
\bibliography{root}

\end{document}